%% file: main.tex
\documentclass{article} 
\usepackage{iclr2020_conference,times}

\input{math_commands.tex}

\usepackage{hyperref}
\usepackage{url}
\usepackage{booktabs}
\usepackage{graphicx}
\usepackage{subcaption}
\title{CTRL: A Conditional Transformer Language Model for Controllable Generation}

\author{Nitish Shirish Keskar\footnotemark[1],\space\space Bryan McCann\thanks{Equal contribution.},\space \space Lav R.\ Varshney,\space\space Caiming Xiong,\space \space Richard Socher\\
Salesforce Research\thanks{{Contact: \tt ctrl-monitoring@salesforce.com}}
}

\iclrfinalcopy 
\begin{document}

\maketitle

\begin{abstract}
Large-scale language models show promising text generation capabilities, but users cannot easily control particular aspects of the generated text.
We release CTRL, a 1.63 billion-parameter conditional transformer language model, trained to condition on control codes that govern style, content, and task-specific behavior.
Control codes were derived from structure that naturally co-occurs with raw text, preserving the advantages of unsupervised learning while providing more explicit control over text generation.
These codes also allow CTRL to predict which parts of the training data are most likely given a sequence. This provides a potential method for analyzing large amounts of data via model-based source attribution.
We have released multiple full-sized, pretrained versions of CTRL at \url{https://github.com/salesforce/ctrl}.
\end{abstract}

\section{Introduction}
With enough data, model capacity, and compute, generative models can learn distributions powerful enough to produce high-quality samples from complex domains. In computer vision, the advent of generative adversarial networks~\citep{goodfellow2014generative} improved image generation. Much research then focused on methods for controlling the generation process and improving estimation of generative distributions~\citep{arjovsky2017wasserstein,chen2016infogan,kingma2013auto}.

In natural language processing, language models are often trained as conditional language models for specific tasks that require text generation~\citep{brants2007large,sutskever2014sequence,rush2015neural}. They are also used as a means of learning word vectors~\citep{mikolov2013distributed}, document vectors~\citep{kiros2015skip}, or contextualized word vectors~\citep{mccann2017learned,peters2018deep,devlin2018bert} for transfer learning. The language models themselves have been transferred to new tasks through fine-tuning as well~\citep{dai2015semi,radford2018improving,howard2018universal}. Less is understood about generation that is not constrained to any specific task. Typically prompts generated by models~\citep{fan2018hierarchical} or written by humans can only be used to provide a rough guide or starting point for the generated text. 
This raises the question of how text generation can be controlled more explicitly.

Inspired by the degree of control available in image generation as well as the recent progress in text generation~\citep{radford2019language} and multitask learning \cite{mccann2018natural}, we train a language model that is conditioned on a variety of control codes~\citep{pfaff1979constraints,poplack1980sometimes} that make desired features of generated text more explicit. With 1.63 billion parameters, our Conditional Transformer Language (CTRL) model can generate text conditioned on control codes that specify domain, style, topics, dates, entities, relationships between entities, plot points, and task-related behavior. 
To preserve the generality of the language model trained in an unsupervised setting, we train CTRL on control codes derived from structure that naturally co-occurs with the raw text typically collected for training large language models. For example, large resources like Wikipedia, Project Gutenberg, and Amazon Reviews can each be assigned a domain-related control code. Smaller resources, like the content extracted from individual subreddits, often occur with both a broader domain name, {\tt reddit}, as well as subdomain information, {\tt r/subdomain}. In the vast majority of cases, text collected for training is associated with a URL, which often contains information pertinent to the text it represents.
Humans can use these codes to trigger generation of text from different linguistic communities without having to understand how to prompt with particular linguistic patterns. Text can be generated in more predictable ways by controlling for content or changing the domain even when the initial prompt remains fixed. 

Because all control codes can be traced back to a particular subset of the training data, CTRL can be used to predict the subset of training data that is most likely given a sequence. This explicit relationship between CTRL and its training data can be exploited to analyze the correlations that the language model has learned from each domain, and it provides a means of studying large amounts of text through the language model.

These control codes also allow for the straightforward inclusion of task-specific data in a way that improves important skills without harming the generality of the model. Control codes for question answering and machine translation make these skills easily accessible with CTRL. These codes can be combined with codes during generation to create novel cross-over between control codes that are task-specific behavior and those that are related to domain and content.

In order to push towards more controllable, general models for natural language processing, we have released multiple full-sized, pretrained versions of CTRL at \url{https://github.com/salesforce/ctrl}. We hope that the release leads to further research into how controllable generation can enhance natural language understanding.

\section{Language Modeling}
\label{sec:background}
Given example sequences of the form $x=(x_1, \ldots, x_n)$ where each $x_i$ comes from a fixed set of symbols, the goal of language modeling is to learn $p(x)$. Because $x$ is a sequence, it is natural to factorize this distribution using the chain rule of probability~\citep{bengio2003neural}:

\[ p(x) = \prod_{i=1}^n p(x_i | x_{<i}) \]

This decomposes language modeling into next-word prediction.
Current state-of-the-art methods~\citep{dai2019transformer,radford2019language} train a neural network with parameters $\theta$ to minimize the negative log-likelihood over a dataset $D=\{x^{1}, \ldots, x^{|D|}\}$: 

\[ \mathcal{L}(D) = - \sum_{k=1}^{|D|} \log {p_{\theta}(x_{i}^{k} | x^{k}_{<i})} \]

Because language models learn ${p_{\theta}(x_i | x_{<i})}$, 
a new $\tilde x$ of length $m$ can be generated by sequentially sampling its constituent symbols: 
$p_{\theta}(x_0), p_{\theta} (x_1 | \tilde x_0), \dots, p_{\theta}(x_m | \tilde x_{<m})$.

\section{Language Modeling with CTRL}
\label{sec:model}
CTRL is a conditional language model that is always conditioned on a control code $c$ and learns the distribution $p(x|c)$.
The distribution can still be decomposed using the chain rule of probability and trained with a loss that takes the control code into account.

\[ p(x|c) = \prod_{i=1}^n p(x_i | x_{<i}, c) \qquad \mathcal{L}(D) = - \sum_{k=1}^{|D|} \log {p_{\theta}(x_{i}^{k} | x^{k}_{<i}, c^k)} \]

The control code $c$ provides a point of control over the generation process. This is true even when sampling $x_0$, in contrast to the traditional language modeling framework described in Sec.~\ref{sec:background}.

CTRL learns $p_{\theta}(x_{i} | x_{<i}, c)$ by training on sequences of raw text prepended with control codes.
After minimal preprocessing (described in Sec.~\ref{sec:experimental_settings}),
a single example sequence containing $n$ tokens is embedded as a sequence of $n$ corresponding vectors in $\mathbb{R}^d$.
Each vector is the sum of a learned token embedding and a sinusoidal positional embedding as in the original Transformer architecture~\citep{vaswani2017attention}.
This sequence of vectors is stacked into a matrix $X_0\in\mathbb{R}^{n \times d}$ so that it can be processed by $l$ attention layers~\citep{vaswani2017attention}.
The $i$th layer consists of two blocks, each of which preserves the model dimension $d$.

The core of the first block is multi-head attention with $k$ heads that uses a causal mask to preclude attending to future tokens:
\begin{align}
\text{Attention}(X, Y, Z)&=\text{softmax}\left(\frac{\text{mask}(XY^\top)}{\sqrt{d}}\right) Z\nonumber\\
\text{MultiHead}(X, k) &= [h_1;\cdots;h_k]W_o\nonumber\\
\text{where } h_j &= \text{Attention}( XW_j^1,  XW_j^2,  XW_j^3) \nonumber
\end{align}

The core of the second block is a feedforward network with ReLU activation~\citep{Nair2010RectifiedLU} that projects inputs to an inner dimension $f$, with parameters $U\in \mathbb{R}^{d \times f}$ and $V\in \mathbb{R}^{f \times d}$:
\begin{align}
FF(X) = \text{max}(0, XU)V \nonumber
\end{align}

Each block precedes core functionality with layer normalization~\citep{Ba2016LayerN, child2019sparse} and follows it with a residual connection~\citep{he2016deep}. Together, they yield $X_{i+1}$:
\begin{align}
 \text{\underline{Block 1}} &&& \text{\underline{Block 2}} \nonumber
\end{align}
\begin{align}
\bar X_i &= \text{LayerNorm}(X_i) & \bar H_i &= \text{LayerNorm}(H_i) \nonumber\\
 H_{i} &= \text{MultiHead}(\bar X_i) + \bar X_i &  X_{i+1} &= \text{FF}(\bar H_i) + \bar H_i\nonumber 
\end{align}

Scores for each token in the vocabulary are computed from the output of the last layer:

\[ \text{Scores}(X_0) = \text{LayerNorm}(X_{l}) W_{vocab} \]

During training, these scores are the inputs of a cross-entropy loss function.
During generation, the scores corresponding to the final token are normalized with a softmax, yielding a distribution for sampling a new token.

\subsection{Data}
\label{sec:data}
We train on $140$ GB of text drawing from a wide variety of domains: Wikipedia (En, De, Es, Fr), Project Gutenberg\footnote{We use a modified version of \url{https://github.com/chiphuyen/lazynlp}}, submissions from 45 subreddits, OpenWebText\footnote{We use a modified version of \url{https://github.com/jcpeterson/openwebtext.git}}, a large collection of news data~\citep{hermann2015teaching,barrault2019findings,sandhaus2008new,N18-1065}, Amazon Reviews~\citep{mcauley2015image}, Europarl and UN data from WMT (En-De, En-Es, En-Fr)~\citep{barrault2019findings}, question-answer pairs (no context documents) from ELI5~\citep{fan2019eli5} and the MRQA shared task\footnote{\url{https://github.com/mrqa/MRQA-Shared-Task-2019}}, which includes the Stanford Question Answering Dataset~\citep{rajpurkar2016squad}, NewsQA~\citep{trischler2016newsqa}, TriviaQA~\citep{joshi2017triviaqa}, SearchQA~\citep{dunn2017searchqa}, HotpotQA~\citep{yang2018hotpotqa}, and Natural Questions~\citep{kwiatkowski2019natural}. 
A full account of training data and associated control codes can be found in Table~\ref{tab:datasource} in the Appendix.

\subsection{Experimental Settings}
\label{sec:experimental_settings}
We learn BPE~\citep{BPE} codes and tokenize the data using fastBPE\footnote{\url{https://github.com/glample/fastBPE}}, but we use a large vocabulary of roughly $250$K tokens.
This includes the sub-word tokens necessary to mitigate problems with rare words, but it also reduces the average number of tokens required to generate long text by including most common words.
We use English Wikipedia and a 5\% split of our collected OpenWebText data for learning BPE codes. 
We also introduce an \texttt{unknown} token so that during preprocessing we can filter out sequences that contain more than $2$ unknown tokens.
This, along with the compressed storage for efficient training (TFRecords)~\citep{tensorflow}, reduces our training data to $140$ GB from the total $180$ GB collected.
Data was treated as a single stream of tokens with non-domain control codes inserted where appropriate (often at document boundaries). 
The stream was chunked into contiguous sequences of tokens.
Each sequence originated from a domain, and it has the corresponding domain control code prepended as the first token in the sequence.
In this way, domain control codes receive special treatment~\citep{kobus2016domain}. 
They are propagated to all text in the domain as the first token. 
This is similar to how codes and natural language sequences have been used in multi-task settings~\citep{wu2016google,johnson2017google,mccann2018natural} to control conditional language models. 
All other control codes are injected into the data without such special treatment~\citep{moryossef2019filling,caswell2019tagged}.
We experimented with sequence lengths of $256$ and $512$ due to memory and optimization constraints.
Despite training on relatively short sequences compared to other approaches, 
we found that a sliding-window approach allows for generation beyond these windows,
and we also found little difference in quality between the two models within the first $256$ tokens. Further, we note that our vocabulary is approximately 4 times larger than similar approaches, hence the effective sequence length in characters is comparable. 

CTRL has model dimension $d=1280$, inner dimension $f=8192$, $48$ layers, and $16$ heads per layer. Dropout with probability $0.1$ follows the residual connections in each layer. Token embeddings were tied with the final output embedding layer \citep{inan,press2016using}. 

CTRL was implemented in TensorFlow~\citep{tensorflow} and trained with a global batch size of $1024$ distributed across $256$ cores of a Cloud TPU v$3$ Pod for $800$k iterations.
Training took approximately 2 weeks using Adagrad \citep{adagrad} with a linear warmup from $0$ to $0.05$ over $25$k steps.
The norm of gradients were clipped to $0.25$ as in \citep{awdlstm}. 
Learning rate decay was not necessary due to the monotonic nature of the Adagrad accumulator. 
We compared to the Adam optimizer~\citep{adam} while training smaller models, 
but we noticed comparable convergence rates and significant memory savings with Adagrad.
We also experimented with explicit memory-saving optimizers including SM3~\citep{sm3}, Adafactor~\citep{adafactor}, and NovoGrad~\citep{novograd} with mixed results. 

\section{Controllable Generation}
\subsection{Sampling}
Typically, temperature-controlled stochastic sampling methods are used for generating text from a trained language model. It is also common to limit the sampling only to the top-$k$ alternatives. Given a temperature $T>0$ and scores $x_i \in \mathbb{R}^d$ for each token $i$ in the vocabulary, the probability of predicting the $i$th token is given by:
\begin{align}
    p_i &= \frac{\exp(x_i/T)}{\sum_j \exp(x_j/T)}.\label{eq:problogits}
\end{align}

The next token is then chosen by sampling through a multinomial distribution with probabilities $p_i$ clipped at the top-$k$ tokens. In the equation above, $T\to0$ approximates a greedy distribution which magnifies the peaks in the probability distribution while $T\to \infty$ flattens the distribution to make it more uniform. Rather than choosing a fixed value of $k$, as is common practice, \citet{nucleus} suggested adapting $k$ heuristically. The nucleus sampling approach chooses a probability threshold $p_t$ and sets $k$ to be the lowest value such that $\sum_i \text{sort}(p_i)>p_t$. If the model is confident in its next-word prediction, then $k$ will be lower and vice versa. Despite the improved generative capabilities of models with such heuristics, there still exists a trade-off between these parameters depending on the generation intended.

\begin{figure}
  \centering
  \begin{subfigure}[b]{0.5\linewidth}
    \centering\includegraphics[width=\columnwidth]{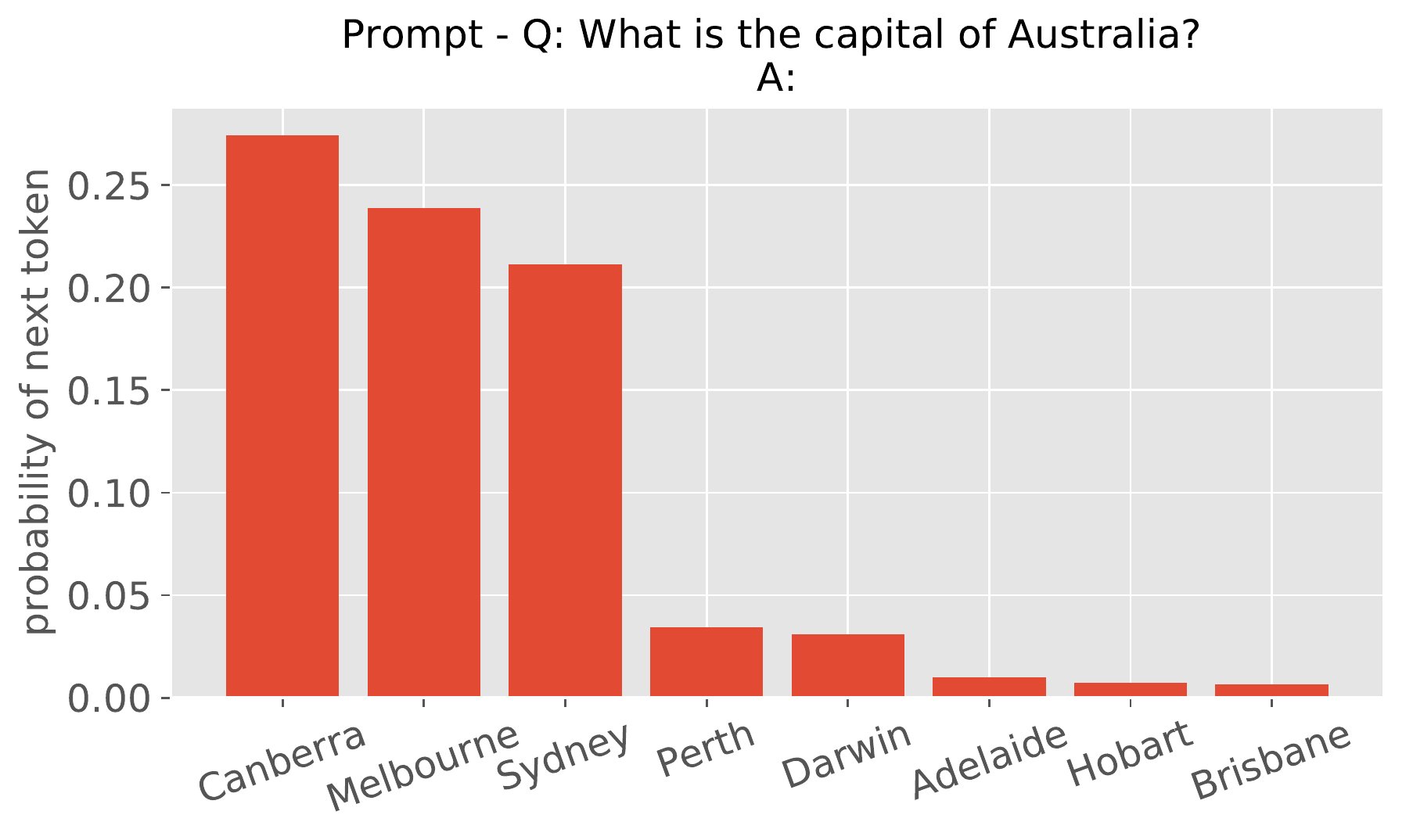}
    \caption{\label{fig:fig1}}
  \end{subfigure}%
  \begin{subfigure}[b]{0.5\linewidth}
    \centering\includegraphics[width=\columnwidth]{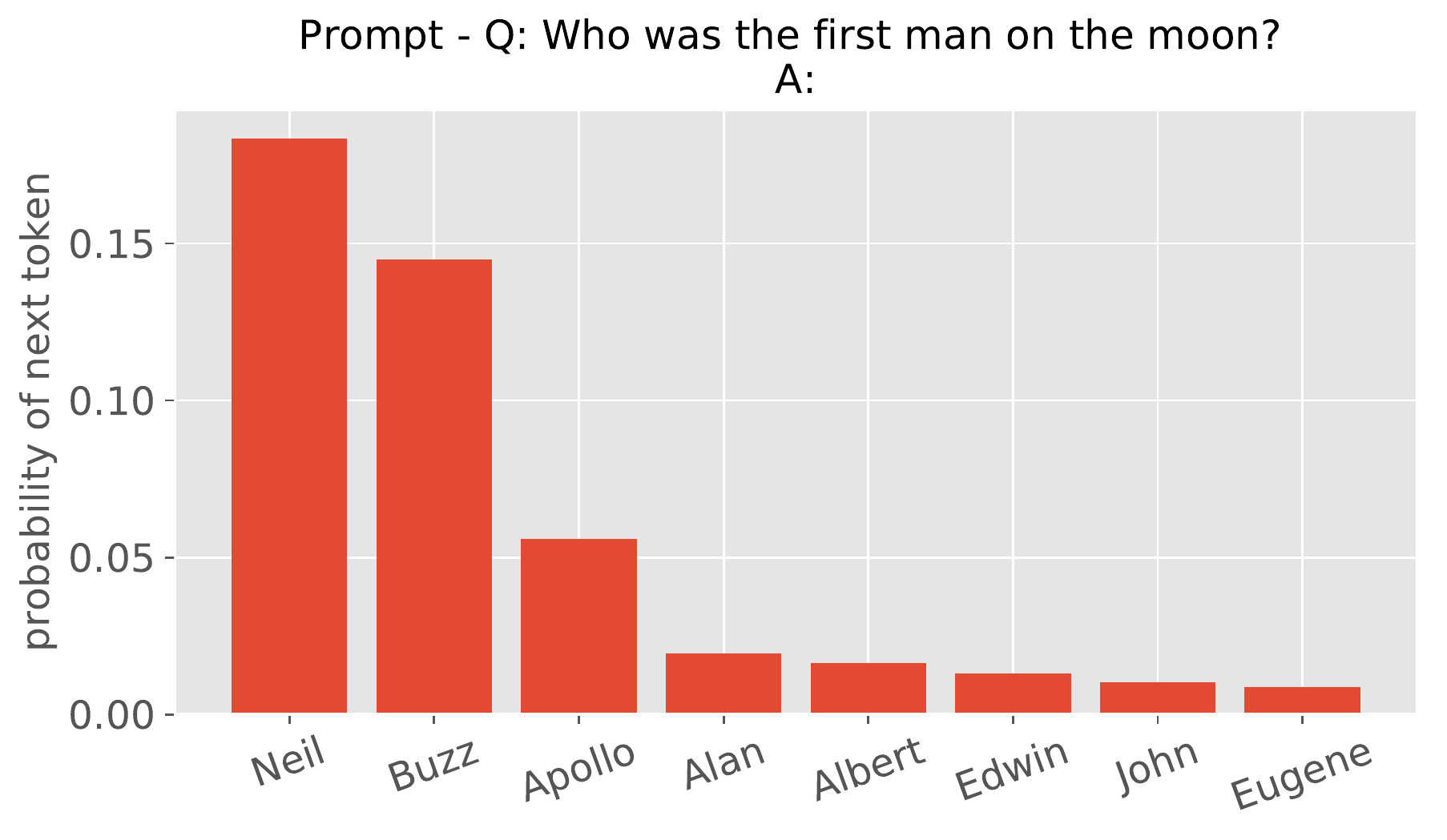}
    \caption{\label{fig:fig2}}
  \end{subfigure}
    \caption{Next-token probability for the prompts \texttt{Q: What is the capital of Australia?} and \texttt{Q: Who was the first man on the moon?} In such cases, sampling using a distribution is detrimental to answering the question correctly.}
    \label{fig:australia}
\end{figure}

Given a prompt: \texttt{Q: What is the capital of Australia?}, a well-trained model assigns higher probability mass to the correct answer, Canberra, but a non-zero probability mass to other cities such as Melbourne, Sydney, Brisbane, Darwin, and Perth, see Figure~\ref{fig:australia}.
By choosing to sample, we mistrust the model, despite it being correct. 
A natural solution to this is to choose the next token greedily. However, this is known to create repetitions of phrases or sentences even for large well-trained models \citep{radford2019language,nucleus}. 
To reconcile the two, we propose a new sampling scheme that trusts the model distribution through near-greedy sampling but prevents repetitions through a penalty. 
This \textit{penalized sampling} works by discounting the scores of previously generated tokens. 
The motivation is similar to coverage mechanisms~\citep{see2017get} and other losses designed to discourage repetition~\citep{unlikelihood}, but penalized sampling is not used during training. 
Given a list of generated tokens $g$, using the notation from \eqref{eq:problogits}, the probability distribution $p_i$ for the next token is defined as:
\begin{align*}
    p_i &= \frac{\exp(x_i/(T \cdot I(i \in g))}{\sum_j \exp(x_j/(T \cdot I(j \in g))} \qquad I(c) = \theta \text{ if c is True else } 1 
\end{align*}

We find that using a greedy sampling and $\theta \approx 1.2$ yields a good balance between truthful generation and lack of repetition. Note that $\theta=1$ is equivalent to \eqref{eq:problogits}. 
We note in passing that this approach succeeds only if the model has learned a sufficiently reliable distribution.

\subsection{Control Codes}
\paragraph{Style by domain.}
\begin{table}
    \centering
    \begin{tabular}{p{13cm}}
\toprule
{\small {\tt \color{red} \small  Wikipedia} {\color{blue} \it \small Anarchism is} a political philosophy that advocates the abolition of all forms of hierarchy and domination, including capitalism, patriarchy, racism, sexism, heterosexism and other oppressive social structures.{\verb \n }{\verb \n }The term "anarchism" was coined by Pierre-Joseph Proudhon in his book "The Poverty of Philosophy" (1844). It has been used to describe various movements within anarchism since then. In its modern sense, it refers to an ideology or movement advocating for social, political, economic and/or cultural change through direct action against existing institutions and practices. Anarchists are often associated with radical left-wing politics, but they have also advocated many different types of libertarian socialism. Some anarchists advocate anarchist communism, others support anarcho-syndicalism, while still others reject both socialism and communism. Many anarchists believe that capitalism should be abolished entirely, though some argue that this would require overthrowing capitalism. Other anarchists oppose any form of state control over private property, such as land ownership, which they see as inherently oppressive. They may favor limited government regulation on certain aspects of life, such as drug use, prostitution, gambling, pornography, abortion, euthanasia, etc.{\verb \n }{\verb \n }History.{\verb \n }{\verb \n }Early history.{\verb \n }{\verb \n }Anarchist ideas were first developed during the French Revolution (1789–1799). During these years, there was widespread opposition to traditional society based upon religion, class, gender, race, ethnicity, nationality, language, education, wealth, and so on. This led to the development of new ideologies like socialism, liberalism, republicanism, communitarianism, individualism/egoism/$\ldots$}\\
      {\small  {\tt \color{red} \small  Books}  {\color{blue} \it \small Anarchism is} the {\verb \n }
 only true and practical form of Socialism. It has been said that Socialism {\verb \n }
 will never be established until it can be proved to work in practice. The {\verb \n }
 same may be said, with equal truth, about Anarchism. But if we are not able {\verb \n }
 to prove its practicability at present, then let us wait for a better time {\verb \n }
 when our present conditions shall have become so much more favorable than {\verb \n }
 they now are, that we shall no longer need any such proof. Let us, then, {\verb \n }
 patiently await this time, which I believe will come sooner or later. In {\verb \n }
 the meantime, as long as there remains one man who believes in it, he must {\verb \n }
 continue to do so. If you wish to see him succeed, try to make him feel his {\verb \n }
 own impotence. You cannot succeed by force alone. He must know himself to {\verb \n }
 be powerless before he begins to work. And what does he do? Why, he goes on {\verb \n }
 working just as hard as ever, because he knows that he is powerless, but {\verb \n }
 still he continues to work, because, after all, he thinks that he is doing {\verb \n }
 right. That is why men like Mr. Owen and myself are called "cranks." We say {\verb \n }
 things that seem absurd to others, but we mean them all the same. When we {\verb \n }
 speak of "the right of labor," we really mean something very different from {\verb \n }
 what most
$\ldots$
}         \\
     {\small     {\tt \color{red} \small  Horror} {\color{blue} \it \small A knife} handle pulled through the open hole in the front. I jumped when the knife hit.{\verb \n }{\verb \n }Eyes widened in horror. Her scream was the only sound I heard besides her sobs.{\verb \n }{\verb \n }The spider touched her feet as it started to dig into the top of her arch. The creature's tears began to flow.{\verb \n }{\verb \n }The spider looked up to her and looked back at me with eyes filled with burning tears. My heart started to race$\ldots$}
         \\
    {\small      {\tt \color{red} \small  Reviews}  {\color{blue} \it \small A knife}
is a tool and this one does the job well.{\verb \n }{\verb \n }Rating: 4.0{\verb \n }{\verb \n }I bought these for my husband who has been using them to cut up his own meat since he got them. He says they are very sharp so be careful when you use them, but that doesn't seem like much of an issue because he's used it on everything from chicken breasts to beef tenderloin$\ldots$}
         \\
    {\small      {\tt \color{red} \small  Relationships}  {\color{blue} \it \small My neighbor is} a jerk and I don't know what to do{\verb \n }{\verb \n }Text: So my neighbors are really nice people. They have been for years. We live in an apartment complex so we get along great.{\verb \n }{\verb \n }But recently they started acting like jerks$\ldots$}
         \\
     {\small     {\tt \color{red} \small  Legal}  {\color{blue} \it \small My neighbor is} threatening to sue me for not letting him use my pool{\verb \n }{\verb \n }Text: I live in a small town and have been renting out the back of my house since last year. The owner has had issues with people using it as well but he never really took any action against them. Last week someone from his property management company came by asking if they could come over because there was an issue with some kids playing on it. He said that we were violating their lease agreement so they would be taking legal action$\ldots$}
         \\
         \bottomrule
    \end{tabular}
    \caption{Even for identical {\color{blue} \it \small prompts}, {\tt \color{red} \small control} {\tt \color{red} \small codes} allow for predictable variation in generation.}
    \label{tab:same_prompt}
\end{table}

Most control codes for our model specify the overall style of generated text by indicating a particular domain of training data.
Examples in Table~\ref{tab:same_prompt} demonstrate that even for identical prompts, control codes allow for predictable variation in generation.
The examples in Table~\ref{tab:more_control} show how CTRL can generate domain-specific text without any prompt.

\begin{table}
    \centering
    \begin{tabular}{p{13cm}}
         \toprule
{\small {\tt \color{red} \small Science  Title:} Scientists have discovered a new type of bacteria that can survive in the presence of high levels of carbon dioxide{\verb \n }Text:[Link to article] (\url{http://news.nationalgeographic.com/news/2011/07/110717-methane-bacteria-carbon-dioxide-environment-science/}){\verb \n }{\verb \n }The researchers found that this particular strain of methane bacterium is able to thrive at concentrations up to 10 times higher than normal atmospheric levels.{\verb \n }{\verb \n }They also note that these findings are consistent with previous research showing that some types of microbes may be capable of surviving under extreme conditions such as those associated with global warming.}\\
{\small {\tt \color{red} \small Politics Title:} The US is the only country in history to have a national debt of more than \textdollar 20 trillion.{\verb \n }Text: I know this isn't exactly an accurate representation but it gives you some idea how much money we are talking about.{\verb \n }{\verb \n }\url{http://en.wikipedia.org/wiki/National_debt#United_States} }\\
{\small {\tt \color{red} \small  Running Text:} I have been running for about a year and a half now but never really got into it.{\verb \n }{\verb \n }
 This past weekend was my first time doing it in the last few months.{\verb \n }{\verb \n }
 My goal is to run at least 3 miles every day this week.}\\
{\small {\tt \color{red} \small  Horror Text:} I was a little girl when my parents got divorced. My dad had been in the military for years and he left me with my mom. She worked as an RN at a hospital so she could take care of me.{\verb \n }{\verb \n }
 When we moved to our new house it took some time before things settled down. We were still living together but there wasn't much going on. It didn't help that my mom would get mad if someone came over or even just walked by her house.{\verb \n }{\verb \n }
 One day while walking through the yard I noticed something out of place...}\\
    {\small      {\tt \color{red} \small  Reviews Rating:} {\tt \color{red} \small 5.0}{\verb \n }{\verb \n }
I have been using this product for a few years and it is the best thing on my hair. It makes your hair soft but not greasy.{\verb \n }{\verb \n }Rating: 5.0{\verb \n }{\verb \n }I've used this product before and love it. It's light weight yet keeps my curls defined without making them crunchy! }\\
    {\small {\tt \color{red} \small  Reviews Rating:} {\tt \color{red} \small 1.0}{\verb \n }{\verb \n }I bought this for my daughter and it was a complete waste of money. The case is very flimsy and the screen protector doesn't fit properly.{\verb \n }{\verb \n }Rating: 1.0{\verb \n }{\verb \n }This product did not work at all. It would charge but when you unplug it, it does nothing. Very disappointed.}\\
         \bottomrule
    \end{tabular}
    \caption{ With CTRL, no {\it \color{blue} \small prompt} is necessary as long as a {\tt \color{red} \small control} {\tt \color{red} \small code} is provided. Control codes can be combined ({\small \tt Reviews}, {\small \tt Rating:}, and {\small \tt VALUE})  to provide finer-grained control. }
    \label{tab:more_control}
\end{table}
\paragraph{More complex control codes.}
Additional control codes can be added to the domain code in order to increasingly constrain generation. 
In Table~\ref{tab:more_control}, adding additional control codes following the domain code further constrains generation. 
These examples demonstrate constraints specifying that the model should start with a title and by specifying a particular rating for reviews. 

Examples of more advanced control are given in Table~\ref{tab:links}. 
In our version of OpenWebText, we include the URL used to downlaod each document as the start of the input sequence.
During training, CTRL learns relationships between the structure of these URLs and the text that follows.
At inference, novel URLs can be used to specify a variety of features: domain, subdomain, entities, entity relations, and even dates. 

\begin{table}
    \centering
    \begin{tabular}{p{13cm}}
         \toprule
{\small {\color{red} \tt \small Links https://www.cnn.com/2007/09/20/us-president-meets-british-pm} 
 LONDON, England (CNN) -- U.S. President George W. Bush met with British Prime Minister Tony Blair on Monday to discuss the war in Iraq, according to a statement from Blair's office.{\verb \n }{\verb \n }The meeting was held at 10 Downing Street and lasted about an hour.{\verb \n }{\verb \n }"The two leaders discussed Iraq and other international issues of mutual concern," said Blair.{\verb \n }{\verb \n }Bush, who is scheduled to meet Wednesday with Russian President Vladimir Putin, will also visit Germany for talks later this week.{\verb \n }{\verb \n }In his statement, Blair said, "We agreed that we should continue our efforts together to bring peace and stability to Iraq. We both reaffirmed our commitment to working closely together, as well as to continuing to work constructively toward achieving lasting security and prosperity throughout the Middle East region."{\verb \n }{\verb \n }Bush's trip comes after he visited Britain last week where he spoke out against terrorism while visiting Buckingham Palace.{\verb \n }{\verb \n }He has been criticized by some lawmakers over what they say are insufficient military resources being devoted to fighting terrorism.}\\
{\small {\tt \color{red} \small  Links https://www.cnn.com/2014/09/20/us-president-meets-british-pm}}\\ {\small JUST WATCHED{\verb \n }{\verb \n }Obama meets with British PM{\verb \n }{\verb \n }MUST WATCH{\verb \n }{\verb \n }Story highlights{\verb \n }{\verb \n }President Barack Obama met with Britain's Prime Minister David Cameron }\\
{\small {\tt \color{red} \small Links https://www.cnn.com/2018/09/20/us-president-meets-british-pm} 
 JUST WATCHED{\verb \n }{\verb \n }Trump and May meet for first time{\verb \n }{\verb \n }MUST WATCH{\verb \n }{\verb \n }Washington (CNN) President Donald Trump, who has been criticized by some in the UK over his decision to leave the European Union, met with British Prime Minister Theresa May, a White House official said on Thursday.}\\
 {\small {\color{red}\tt \small  Links https://www.cnn.com/09/20/2018/star-spotted} 
JUST WATCHED{\verb \n }{\verb \n }Star spotted: The world's most elusive star{\verb \n }{\verb \n }MUST WATCH{\verb \n }{\verb \n }Story highlights{\verb \n }{\verb \n }A new study suggests that the Milky Way is home to a supermassive black hole{\verb \n }{\verb \n }It may be one of our closest galactic neighbors{\verb \n }{\verb \n }(CNN) Astronomers have discovered what they believe could be an extremely rare astronomical event: An object so bright it can only be seen with powerful telescopes.{\verb \n }{\verb \n }This object, dubbed SN2014J, was first detected by NASA's Spitzer Space Telescope in 2014. $\ldots$}\\
{\small {\color{red}\tt \small  Links https://www.etonline.com/09/20/2018/star-spotted}{\verb \n }{\verb \n }EXCLUSIVE: Star Spotted in New York City{\verb \n }{\verb \n }By Kelly Marot | September 20, 2018 10:51 AM PDT{\verb \n }{\verb \n }Getty Images{\verb \n }{\verb \n }The star of the new season of “Stranger Things” is making a splash on her own turf — and it’s not even Halloween.{\verb \n }{\verb \n }Winona Ryder has been seen out and about with friends at night time in Manhattan this week, E! News can exclusively reveal.{\verb \n }{\verb \n }A source tells us, “She was hanging around with some friends, but she wasn’t doing anything too crazy. She just wanted to get away from it all for a little bit.”$\ldots$}\\
{\small {\color{red}\tt \small  Links https://www.cnn.com/style/09/20/2018/george-clooney-interview} George Clooney on the future of his acting career{\verb \n }{\verb \n }By{\verb \n }{\verb \n }Updated 10:51 AM ET, Thu September 20, 2018{\verb \n }{\verb \n }Chat with us in Facebook Messenger. Find out what's happening in the world as it unfolds.{\verb \n }{\verb \n }Photos:George Clooney, 'Ocean's 8'{\verb \n }{\verb \n }Actor George Clooney attends a photocall for "Ocean's 8" at Grauman's Chinese Theatre on August 31, 2018, in Los Angeles.{\verb \n }{\verb \n }$\ldots$}\\
 {\small {\color{red}\tt \small  Links https://www.cnn.com/politics/09/20/2018/george-clooney-interview} JUST WATCHED{\verb \n }{\verb \n }George Clooney on the Trump administration{\verb \n }{\verb \n }MUST WATCH{\verb \n }{\verb \n }(CNN) Actor and activist George Clooney, who has been a vocal critic of President Donald Trump, said he is "ready to go back into the political arena" after his role in an anti-Trump documentary was cut from theaters this week.{\verb \n }{\verb \n }$\ldots$}
 \\
         \bottomrule
    \end{tabular}
    \caption{CTRL is trained with links as {\tt \color{red} \small  control} {\tt \color{red} \small  codes}. Links provide a way to specify domain, subdomain, entities, entity relations, and even date. The links in these examples do not actually link to text; users can mimic the structure of the URLs that appear during training to create novel content during generation. Note that {\tt us-president} is interpreted differently by the model depending on the date used ({\tt 2007},{\tt  2014},{\tt  2018)}. Similarly, {\tt star} is interpreted differently based on the domain ({\tt cnn} vs. {\tt etonline}) and topic ({\tt style} vs. {\tt politics}) can be varied even for identical entities ({\tt george-clooney}).}
    \label{tab:links}
\end{table}

\paragraph{Triggering specific tasks.}
\begin{table}
    \centering
    \begin{tabular}{p{13cm}}
\toprule
{\small {\tt \color{red} \small  Questions} {\color{blue} \it \small Q: What is the capital of India?} {\small {\tt \color{red} \small  A:}} {\small
New Delhi{\verb \n }{\verb \n }Q: Which city in India has a large population of people from Gujarat?{\verb \n }{\verb \n }A: Ahmedabad{\verb \n }{\verb \n }Q: How many languages are spoken by Indians living abroad?{\verb \n }{\verb \n }A: more than 100$\ldots$}}\\
      {\small  
      {\tt \color{red} \small  Translation}  {\color{red} \tt \small English} {\color{red} \tt \small:} {\color{blue} \it \small We release a new model for coherent language generation} {\color{red} \tt \small;} {\color{red} \tt \small French} {\color{red} \tt \small:} Nous publions un nouveau mod\'ele de g\'en\'eration coh\'erente du langage{\verb \n }{\verb \n }English : The European Parliament has just adopted the report on the proposal for a regulation of the European Parliament and of the Council establishing an action programme to promote multilingualism in Europe. ; French : Le Parlement europ\'een vient d'adopter le rapport sur la proposition de r\`eglement du Parlement et du Conseil \'etablissant un programme-cadre pour promouvoir le pluriling\:uisme dans l'Union europ\'eenne$\ldots$}
         \\
         \bottomrule
    \end{tabular}
    \caption{More complex templatized control codes are used for task-specific generation.}
    \label{tab:tasks}
\end{table}
A small number of control codes are related to specific tasks like question answering and translation. These codes  constrain the generation process the most, by triggering task-specific generation. In Table~\ref{tab:tasks}, we demonstrate relatively complex control codes for question answering and machine translation that act as a template mixed with a natural language prompt. 

\paragraph{Zero-shot code-mixing.}
\begin{table}
    \centering
    \begin{tabular}{p{13cm}}
\toprule
{\small {\tt \color{red} \small  Diet  English} {\tt \color{red} \small :} {\color{blue} \it \small I lost 10 kgs!} {\tt \color{red} \small ;} {\tt \color{red} \small  German} {\tt \color{red} \small  :} Ich habe 10 Kilogramm verloren!{\verb \n }{\verb \n }English : It is a great achievement for me and my family. ; German : Das ist eine große Leistung für mich und meine Familie.{\verb \n }{\verb \n }English : Thank you very much to all the people who have helped me. ; German : Vielen Dank an alle, die mir geholfen haben.$\ldots$}\\
      {\small  {\tt \color{red} \small  Politics  Title:}  {\color{blue} \it \small Les Etats-Unis sont un}  pays de droite{\verb \n }{\verb \n }Text: Je suis Am\'ericain et je ne peux pas supporter ce que font les USA en mati\`ere d'immigration{\verb \n }{\verb \n }J'ai d\'ej\`a \'ecrit sur le sujet mais il n'y a aucune r\'eaction \url{http://www.reddit.com/r/politics/comments/1tqjzw/les_etats_unes_son_tres-de-gauche/} $\ldots$}
         \\
         \bottomrule
    \end{tabular}
    \caption{Some codes can be mixed to generate text with novel cross-over behavior. In Table~\ref{tab:mixing}, we present two examples. In the first example, we mix translation codes into the {\tt Diet} domain. By doing so, the model continues alternatively generates English and German sentences while respecting the {\tt Diet} domain and remains coherent across translations. In the second example, the {\tt Politics} domain is mixed with a French prompt despite never seeing this combination in training. }
    \label{tab:mixing}
\end{table}
In the first example we mix a diet subreddit (r/keto) with machine translation control codes for English and German. In contrast to using {\small \tt Translation } in~\ref{tab:more_control}, the generated text with mixed codes is coherent across multiple translated lines. This structure is an influence of {\small \tt Diet} because it had multiline examples in the training data, whereas the translation data consisted of shuffled single lines. In the second example we mix the politics subreddit (r/politics) with a prompt that starts in French though no examples of this kind were found in the training data.

\section{Source Attribution}
{ 
\renewcommand{\arraystretch}{1.1}
\begin{table}
    \centering
    \begin{tabular}{p{7cm}l}
    \toprule
         Query Prompt &  Attributed Sources\\
        \midrule
        {\small Global warming is a lie.} 
        & 
        {\small r/unpopularopinion, r/conspiracy, r/science}\\
        
        {\small Global warming is a lie }
        & 
        {\small r/eli5, r/science, r/unpopularopinion}\\
        {\small Global warming is a real phenomenon} 
        &
        {\small  r/eli5, r/science, r/changemyview}\\
        {\small Global warming is a real phenomenon.}
        &
        {\small  OpenWebText, r/changemyview, r/science} \\
        {\small  I don't think women should be allowed to vote.} 
        &
        {\small  r/christianity, r/atheism, r/unpopularopinion} \\
        {\small Carbs are your enemy when you want to get lean.}
        &
        {\small  r/fitness, r/loseit, r/keto }\\
        {\small I just want to be a fun aunt. I'm not interested in babies.}& {\small r/babybumps, r/childfree, r/twoxchromosome}\\
        {\small My landlord is suing me for unpaid rent.} & {\small r/legaladvice, r/personalfinance, r/frugal}\\
        {\small FROM fairest creatures we desire increase,{\verb \n }{\verb \n }That thereby beauty's rose might never die }& {\small Gutenberg, Wikipedia, OpenWebText} \\
            \bottomrule
    \end{tabular}
    \caption{We probe CTRL for learned correlations between sequences and domains. Note that this procedure is sensitive to small changes in the prompt. For example, "Global warming is a lie" differs from "Global warming is a lie." r/eli5 stands for "Explain like I'm five". Attribution experiments use the model trained on sequences of length $256$; it was trained longer and provided better estimation of source. Source attribution cannot be considered a measure of veracity, but only a measure of how much each domain token influences a given sequence.}
    \label{tab:sourcing}
\end{table}
}
The domain control codes can be used to partition the training data into mutually exclusive sets. This supports a simple method for determining which subsets of the training data the language model considers most likely given a sequence. Recall that the language model has learned a distribution $p_{\theta}(x|c)$. By specifying a prior over domain control codes for $p(c)$, it is straightforward to compute a ranking of domains:
\[ p_{\theta}(c|x) \propto p_{\theta}(x|c) p(c) \]

We found that the empirical prior of the training data weights domains with large amounts of data too heavily. Instead, we use a uniform prior over the domain control codes. Examples can be found in Table~\ref{tab:sourcing}.

We note that the data used to train this model does not have universal coverage and contains the cultural associations present in the original sources. All applications of the model inherently depend on those original associations for prediction. In fact, this method of source attribution relies on exploiting the original associations to establish relationships between the language model and its training data.

The model does not have a notion of whether any particular cultural association is good or bad, right or wrong, true or false. It only learns correlations between cultural associations and domains. This is evidenced by the fact that contradictory statements are often attributed to the same sources: competing claims often appear in the same contexts. CTRL provides model-based evidence that certain domains are more likely to contain language similar to given statements, but it should not be used to make normative or prescriptive claims. It is a descriptive tool for analyzing correlations in large amounts of text.  

 \section{Related Work}
 \paragraph{Language modeling.}
 Language models~\citep{bengio2003neural} have played an important role in natural language processing through transferrable word vectors~\citep{mikolov2013distributed}, contextualized word vectors~\citep{peters2018deep,devlin2018bert,lample2019cross}, and models~\citep{howard2018universal,radford2018improving}.
 Recent work on memory mechanisms~\citep{dai2019transformer,lample2019large} has improved perplexities on the most common benchmarks, and even without these memories, large Transformer architectures~\citep{vaswani2017attention} like GPT-2~\citep{radford2019language}, OpenGPT-2\footnote{{ \scriptsize \url{https://blog.usejournal.com/opengpt-2-we-replicated-gpt-2-because-you-can-too-45e34e6d36dc}}}, and Megatron\footnote{{ \scriptsize\url{https://github.com/NVIDIA/Megatron-LM}}} can achieve state-of-the-art results without directly training for any particular language modeling benchmark. Because these latter language models are trained on far more diverse data than is used in the supervised setting, they demonstrate impressive text generation capabilities~\citep{radford2019language,zellers2019defending}.
 
 \paragraph{Multi-task learning.}
 These models demonstrate the potential to learn multiple tasks as well as quick adaptation to patterns in input prompts~\citep{radford2019language}.
 This potential showed that language models can offer an alternative to supervised multi-task learning as framed by several recent benchmarks~\citep{wang2018glue,mccann2018natural}.
 Language models might also offer a foundation to extend proposals of unified, multi-task systems for all of NLP~\citep{collobert2008unified,collobert2011natural}, parsing and tagging~\citep{hashimoto2016joint}, multiple languages~\citep{wu2016google,johnson2017google}, and multiple modalities~\citep{luong2015multi,kaiser2017one}. Several works have pointed to natural language as a means for controlling these multi-task systems~\citep{mccann2018natural,radford2019language,keskar2019unifying}, and several point to the benefits of a code book either specified explicitly~\citep{wu2016google} or learned in a latent space~\citep{kaiser2018fast}. This work attempts to balance these approaches.
 
 \paragraph{Sampling methods and coverage mechanisms.}
 Recent work in sampling methods for text generation has focused on reducing repetition by replacing it with novel, coherent text~\citep{fan2018hierarchical,nucleus}. The problem of repetition can instead be approached by altering the training objectives, as with coverage mechanisms~\citep{see2017get} and context-based losses~\citep{unlikelihood}. When prioritizing control, the trade-off between novelty in the generated text and consistency with prompts and prior generated text remains a difficult challenge, but this work found that relying on inference-time methods~\citep{fan2018hierarchical,nucleus} that are closer in behavior to context-based losses~\citep{see2017get,unlikelihood} provides a reasonable solution as long as the distribution of the language model is sufficiently confident in its decisions.

\section{Future Directions}
\paragraph{More control codes and finer-grained control.}
The particular choice of control codes in this work is intended to represent a reasonably large variety in control over domain, topic, entities, entity relations, and dates. 
A very flexible means of control is through the natural structure of the internet in the form of URLs. 
Many of the domains that were mapped in this work to a single control code (e.g. Wikipedia, Project Gutenberg), could be refined to provide more fine-grained control either through further exploitation of URL structure (\url{en.wikipedia.org}, \url{de.wikipedia.org}, \url{en.wikipedia.org/wiki/Anarchism}, \url{en.wikipedia.org/wiki/Anarchism#History}) or through the manual extraction of structure already present in the data (e.g. {\tt Books Author Title Chapter}). 
We hope future work explores extensions of CTRL to new domains in ways that provide further insight into controllable text generation.

\paragraph{Extensions to other areas in NLP.}
This work suggests that including data for specific tasks need not harm the general nature of an unsupervised learning process. For important skills, the inclusion of supervised data or task-specific data generated through unsupervised means~\citep{artetxe2017unsupervised,lewis2019unsupervised} can lead to obvious improvements. While this work experimented with trivia-style question answering (without context documents) and small amounts of machine translation data, it remains an open question whether these language models can learn to effectively perform tasks like extractive question answering or state-of-the-art multilingual machine translation while still preserving general pattern recognition and text generation functionality. 

Many tasks present difficult challenges to the supervised setting. Commonsense reasoning \citep{levesque2012winograd} and abstractive summarization~\citep{rush2015neural} represent two areas where these challenges remain readily apparent~\citep{kryscinski2019neural}. Yet language models show potential for mitigating these problems directly~\citep{trinh2018simple,radford2019language} or indirectly~\citep{rajani2019explain,xenouleas2019sumqe,scialom2019answers}. We hope that in future work CTRL can be extended to far more tasks through the use of both unsupervised and supervised techniques. 

\paragraph{Analyzing the relationships between language models and training data.}
CTRL is trained on a small subset of the possible data available. Therefore the model is biased towards the patterns of language used in the training data. The data is likely not representative of many linguistic communities, but CTRL offers an explicit method for analyzing the relationship between the model and its current training data. As methods improve, more data is collected, and training of these large models continues, we hope to use this tool to better understand the particular cultural associations the model learns from each data source.  

\paragraph{Making the interface between humans and language models more explicit and intuitive.}
CTRL is designed to make the interface between humans and language models more intuitive. Text generation can be a powerful tool for enhancing creativity and exploration. In future work, we hope to study how the beneficial applications of such models can be enhanced by providing more control to human users.

\section{CTRL-ALT-DEL: The Ethics of Large Language Models}
Openness and replicability are central aspects of the scientific ethos that, prima facie, suggest the release of complete scientific research results. We reify these principles by releasing all trained CTRL models. 

Although much scientific research and innovation can benefit the public, it may also be diverted to harmful uses or have unintended negative impacts (without animus).  
\citet{Brundage_ea2019}, among others, have argued artificial intelligence has such an omni-use character and have suggested governance policies emerging from the \emph{responsible innovation} literature \citep{Brundage2016}.  
Historical evidence has pointed to the inadequacy of self-moratoriums for governing omni-use technologies \citep{KaiserM2012}; we take a course of action that differs from such self-regulation. 
Our actions reflect principles from a recent sociology-based AI governance framework that aims to expand responsible innovation to consider networks of users, dynamics, and feedback  \citep{VarshneyKS2019}.  

\begin{itemize}
    \item Rather than self-governance, we sought to diversify inputs to governance through pre-release review from experts at the Partnership on AI (PAI).  These experts, in turn, drew on emerging norms and governance processes that incorporate a broad set of values from across society.
    
    \item Prior to release, the research team conducted a technology foresight exercise to anticipate possible malicious use cases.  In particular, we used a scenario planning approach to technology foresight that systematically attempts to envision plausible longer-term future states of science, technology, and society.  This anticipatory focus on possibilities rather than probabilities lessens several shortcomings of formal risk assessment in the face of contested assumptions, which has proven ineffective in identifying the most profound future impacts of innovation \citep{StilgoeOM2013}.
    
    \item As part of our model release, we include a code of conduct in the README at \url{https://github.com/salesforce/ctrl}. 
    This code of conduct is modeled after emerging community norms ensconced in the Do No Harm and Just World Licenses. Simultaneously recognizing that it has no legal force and that users are agents of technological change embedded in social networks, the aim is to encourage reflection at the consumption junction \citep{Cowan1987} through norm-setting and reduce unintended uses.

    \item The README also includes a subset of the questions that the team discussed when deliberating release of the models, drawn from early drafts of community-driven PAI documents (to be released in the near future). This may further encourage users to reflect on norms and responsibilities associated with models that generate artificial content. In particular, users are asked to share answers to the included questions, to pose further questions, and suggest solutions by emailing \url{ctrl-monitoring@salesforce.com}.
    
    \item Finally, the README asks users to develop appropriate documentation \citep{PAI_aboutML,Arnold_ea2018,MitchellWZBVHSRG2019} when building on CTRL and to tell the research team how they are using CTRL by emailing \url{ctrl-monitoring@salesforce.com}. This facilitates a post-release monitoring plan that observes how people are using CTRL in the wild (together with active observations).  Such \emph{post-market} plans recognize that most innovations are unexpected and hard to forecast.  It is intended to enable a responsive approach to responsible innovation, not just with respect to harmful uses but also unintended negative impacts without animus.
\end{itemize}

\section{Conclusion}

With 1.63 billion parameters, CTRL is the largest publicly released language model to date. It is trained with control codes so that text generation can be more easily controlled by human users. These codes allow users to explicitly specify domain, subdomain, entities, relationships between entities, dates, and task-specific behavior. We hope that the release of this model at \url{https://github.com/salesforce/ctrl} pushes towards more controllable, general models for natural language processing, and we encourage future discussion about artificial generation with our team by emailing \url{ctrl-monitoring@salesforce.com}. 

\section{Acknowledgements}
We would like to thank Kathy Baxter for her help in the ethical considerations of our work and facilitating the external review process; Srinath Meadusani, Lavanya Karanam, Ning Dong, and Navin Ramineni for their help with setting up and maintaining compute infrastructure; Zak Stone and his team at Google for assistance with TPU infrastructure and code; and Joseph Olsen, Roy Davis, Joshua Simmons, Denise Lo, and Sam Edwards for their help with open sourcing.

\bibliography{iclr2020_conference}
\bibliographystyle{iclr2020_conference}

\clearpage
\appendix
\section{Data sources and breakdown}
\begin{table}[h]
    \small
    \centering
    \begin{tabular}{ll}
    Control Code & Description \\
    \toprule
    Wikipedia & English Wikipedia\\
    Books & Books from Project Gutenberg \\
    Reviews & Amazon Reviews data~\citep{mcauley2015image}\\
    Links & OpenWebText (See Sec.~\ref{sec:experimental_settings})\\
    Translation & WMT translation date \citep{barrault2019findings}\\
    News & News articles from CNN/DailyMail~\cite{nallapati2016abstractive}, New York Times \\ & and Newsroom~\citep{N18-1065} \\
    multilingual & Wikipedias in German, Spanish and French\\
    Questions & (Questions and answers only) MRQA shared task (See Section~\ref{sec:data})\\
    Explain & (Only main post) \citep{fan2019eli5}\\
    \midrule
    \multicolumn{2}{c}{Sub-reddit data (Title, Text and Score/Karma) collected from \url{pushshift.io}. }\\
    \midrule
    Alone & \texttt{r/childfree}\\
    Atheism & \texttt{r/atheism}\\
Christianity & \texttt{r/christianity}\\
Computing & \texttt{r/computing}\\
Confession & \texttt{r/offmychest}\\
Confessions & \texttt{r/confession}\\
Conspiracy & \texttt{r/conspiracy}\\
Diet & \texttt{r/keto}\\
Extract & \texttt{r/childfree}\\
Feminism & \texttt{r/twoxchromosome}\\
Finance & \texttt{r/personalfinance}\\
Fitness & \texttt{r/fitness}\\
Funny & \texttt{r/funny}\\
Gaming & \texttt{r/gaming}\\
Horror & \texttt{r/nosleep}\\
Human & \texttt{r/nfy}\\
India & \texttt{r/india}\\
Joke & \texttt{r/jokes}\\
Joker & \texttt{r/joke}\\
Learned & \texttt{r/todayilearned}\\
Legal & \texttt{r/legaladvice}\\
Movies & \texttt{r/movies}\\
Netflix & \texttt{r/netflix}\\
Norman & \texttt{r/lifeofnorman}\\
Notion & \texttt{r/unpopularopinion}\\
Opinion & \texttt{r/changemyview}\\
Politics & \texttt{r/politics}\\
Pregnancy & \texttt{r/babybumps}\\

Relationship & \texttt{r/relationshipadvice}\\
Relationships & \texttt{r/relationships}\\
Retail & \texttt{r/talesfromretail}\\
Running & \texttt{r/running}\\
Saving & \texttt{r/frugal}\\
Scary & \texttt{r/scaryshortstories}\\
Science & \texttt{r/science}\\
Technologies & \texttt{r/technology}\\
Teenage & \texttt{r/teenager}\\
Thoughts & \texttt{r/showerthoughts}\\
Tip & \texttt{r/lifeprotips}\\
Weight & \texttt{r/loseit}\\
Writing & \texttt{r/writingprompts}\\
\bottomrule
    \end{tabular}
    \caption{Data and control codes. Wikipedia, Books, News and multilingual have no secondary code. {\tt Reviews} can be followed by {\tt Rating:} and a value of \{{\tt 1.0, 2.0, 3.0, 4.0, 5.0}\}. For Links, a full or partial URL can be provided (See Table~\ref{tab:links}). For all the Reddit data, the secondary code can be \texttt{Title:} or \texttt{Text:}, which is the title and text of the article, respectively.}
    \label{tab:datasource}
\end{table}
\end{document}

%% file: math_commands.tex
\usepackage{amsmath,amsfonts,bm}

\def\eqref#1{equation~\ref{#1}}

\def\1{\bm{1}}

\DeclareMathAlphabet{\mathsfit}{\encodingdefault}{\sfdefault}{m}{sl}
\SetMathAlphabet{\mathsfit}{bold}{\encodingdefault}{\sfdefault}{bx}{n}













%% file: main.bbl
\begin{thebibliography}{3}
\providecommand{\natexlab}[1]{#1}
\providecommand{\url}[1]{\texttt{#1}}
\expandafter\ifx\csname urlstyle\endcsname\relax
  \providecommand{\doi}[1]{doi: #1}\else
  \providecommand{\doi}{doi: \begingroup \urlstyle{rm}\Url}\fi

\bibitem[Bengio \& LeCun(2007)Bengio and LeCun]{Bengio+chapter2007}
Yoshua Bengio and Yann LeCun.
\newblock Scaling learning algorithms towards {AI}.
\newblock In \emph{Large Scale Kernel Machines}. MIT Press, 2007.

\bibitem[Goodfellow et~al.(2016)Goodfellow, Bengio, Courville, and
  Bengio]{goodfellow2016deep}
Ian Goodfellow, Yoshua Bengio, Aaron Courville, and Yoshua Bengio.
\newblock \emph{Deep learning}, volume~1.
\newblock MIT Press, 2016.

\bibitem[Hinton et~al.(2006)Hinton, Osindero, and Teh]{Hinton06}
Geoffrey~E. Hinton, Simon Osindero, and Yee~Whye Teh.
\newblock A fast learning algorithm for deep belief nets.
\newblock \emph{Neural Computation}, 18:\penalty0 1527--1554, 2006.

\end{thebibliography}


\begin{thebibliography}{86}
\providecommand{\natexlab}[1]{#1}
\providecommand{\url}[1]{\texttt{#1}}
\expandafter\ifx\csname urlstyle\endcsname\relax
  \providecommand{\doi}[1]{doi: #1}\else
  \providecommand{\doi}{doi: \begingroup \urlstyle{rm}\Url}\fi

\bibitem[PAI(2019)]{PAI_aboutML}
Annotation and benchmarking on understanding and transparency of machine
  learning lifecycles {(ABOUT ML)}, 2019.
\newblock URL \url{https://www.partnershiponai.org/about-ml/}.
\newblock Partnership on AI (PAI), v0.

\bibitem[Abadi et~al.(2016)Abadi, Barham, Chen, Chen, Davis, Dean, Devin,
  Ghemawat, Irving, Isard, et~al.]{tensorflow}
Mart{\'\i}n Abadi, Paul Barham, Jianmin Chen, Zhifeng Chen, Andy Davis, Jeffrey
  Dean, Matthieu Devin, Sanjay Ghemawat, Geoffrey Irving, Michael Isard, et~al.
\newblock Tensorflow: A system for large-scale machine learning.
\newblock In \emph{12th $\{$USENIX$\}$ Symposium on Operating Systems Design
  and Implementation ($\{$OSDI$\}$ 16)}, pp.\  265--283, 2016.

\bibitem[Anil et~al.(2019)Anil, Gupta, Koren, and Singer]{sm3}
Rohan Anil, Vineet Gupta, Tomer Koren, and Yoram Singer.
\newblock Memory-efficient adaptive optimization for large-scale learning.
\newblock \emph{arXiv preprint arXiv:1901.11150}, 2019.

\bibitem[Arjovsky et~al.(2017)Arjovsky, Chintala, and
  Bottou]{arjovsky2017wasserstein}
Martin Arjovsky, Soumith Chintala, and L{\'e}on Bottou.
\newblock Wasserstein generative adversarial networks.
\newblock In \emph{International conference on machine learning}, pp.\
  214--223, 2017.

\bibitem[Arnold et~al.(2018)Arnold, Bellamy, Hind, Houde, Mehta, Mojsilovic,
  Nair, Natesan~Ramamurthy, Reimer, Olteanu, Piorkowski, Tsay, and
  Varshney]{Arnold_ea2018}
Matthew Arnold, Rachel K.~E. Bellamy, Michael Hind, Stephanie Houde, Sameep
  Mehta, Aleksandra Mojsilovic, Ravi Nair, Karthikeyan Natesan~Ramamurthy,
  Darrell Reimer, Alexandra Olteanu, David Piorkowski, Jason Tsay, and Kush~R.
  Varshney.
\newblock Factsheets: Increasing trust in {AI} services through supplier's
  declarations of conformity, August 2018.
\newblock arXiv:1808.07261 [cs.CY].

\bibitem[Artetxe et~al.(2017)Artetxe, Labaka, Agirre, and
  Cho]{artetxe2017unsupervised}
Mikel Artetxe, Gorka Labaka, Eneko Agirre, and Kyunghyun Cho.
\newblock Unsupervised neural machine translation.
\newblock \emph{arXiv preprint arXiv:1710.11041}, 2017.

\bibitem[Ba et~al.(2016)Ba, Kiros, and Hinton]{Ba2016LayerN}
Jimmy Ba, Ryan Kiros, and Geoffrey~E. Hinton.
\newblock Layer normalization.
\newblock \emph{CoRR}, abs/1607.06450, 2016.

\bibitem[Barrault et~al.(2019)Barrault, Bojar, Costa-juss{\`a}, Federmann,
  Fishel, Graham, Haddow, Huck, Koehn, Malmasi, et~al.]{barrault2019findings}
Lo{\"\i}c Barrault, Ond{\v{r}}ej Bojar, Marta~R Costa-juss{\`a}, Christian
  Federmann, Mark Fishel, Yvette Graham, Barry Haddow, Matthias Huck, Philipp
  Koehn, Shervin Malmasi, et~al.
\newblock Findings of the 2019 conference on machine translation (wmt19).
\newblock In \emph{Proceedings of the Fourth Conference on Machine Translation
  (Volume 2: Shared Task Papers, Day 1)}, pp.\  1--61, 2019.

\bibitem[Bengio et~al.(2003)Bengio, Ducharme, Vincent, and
  Jauvin]{bengio2003neural}
Yoshua Bengio, R{\'e}jean Ducharme, Pascal Vincent, and Christian Jauvin.
\newblock A neural probabilistic language model.
\newblock \emph{Journal of machine learning research}, 3\penalty0
  (Feb):\penalty0 1137--1155, 2003.

\bibitem[Brants et~al.(2007)Brants, Popat, Xu, Och, and Dean]{brants2007large}
Thorsten Brants, Ashok~C Popat, Peng Xu, Franz~J Och, and Jeffrey Dean.
\newblock Large language models in machine translation.
\newblock In \emph{Proceedings of the 2007 Joint Conference on Empirical
  Methods in Natural Language Processing and Computational Natural Language
  Learning (EMNLP-CoNLL)}, pp.\  858--867, 2007.

\bibitem[Brundage(2016)]{Brundage2016}
Miles Brundage.
\newblock Artificial intelligence and responsible innovation.
\newblock In Vincent~C. M{\"{u}}ller (ed.), \emph{Fundamental Issues of
  Artificial Intelligence}, pp.\  543--554. Springer, 2016.

\bibitem[Brundage et~al.(2019)Brundage, Avin, Clark, Toner, Eckersley,
  Garfinkel, Dafoe, Scharre, Zeitzoff, Filar, Anderson, Roff, Allen,
  Steinhardt, Flynn, {\'{O}}~{h\'{E}igeartaigh}, Beard, Belfield, Farquhar,
  Lyle, Crootof, Evans, Page, Bryson, Yampolskiy, and Amodei]{Brundage_ea2019}
Miles Brundage, Shahar Avin, Jack Clark, Helen Toner, Peter Eckersley, Ben
  Garfinkel, Allan Dafoe, Paul Scharre, Thomas Zeitzoff, Bobby Filar, Hyrum
  Anderson, Heather Roff, Gregory~C. Allen, Jacob Steinhardt, Carrick Flynn,
  Se{\'{a}}n {\'{O}}~{h\'{E}igeartaigh}, Simon Beard, Haydn Belfield, Sebastian
  Farquhar, Clare Lyle, Rebecca Crootof, Owain Evans, Michael Page, Joanna
  Bryson, Roman Yampolskiy, and Dario Amodei.
\newblock The malicious use of artificial intelligence: Forecasting,
  prevention, and mitigation, February 2019.
\newblock arXiv:1802.07228 [cs.AI].

\bibitem[Caswell et~al.(2019)Caswell, Chelba, and Grangier]{caswell2019tagged}
Isaac Caswell, Ciprian Chelba, and David Grangier.
\newblock Tagged back-translation.
\newblock \emph{arXiv preprint arXiv:1906.06442}, 2019.

\bibitem[Chen et~al.(2016)Chen, Duan, Houthooft, Schulman, Sutskever, and
  Abbeel]{chen2016infogan}
Xi~Chen, Yan Duan, Rein Houthooft, John Schulman, Ilya Sutskever, and Pieter
  Abbeel.
\newblock Infogan: Interpretable representation learning by information
  maximizing generative adversarial nets.
\newblock In \emph{Advances in neural information processing systems}, pp.\
  2172--2180, 2016.

\bibitem[Child et~al.(2019)Child, Gray, Radford, and
  Sutskever]{child2019sparse}
Rewon Child, Scott Gray, Alec Radford, and Ilya Sutskever.
\newblock Generating long sequences with sparse transformers.
\newblock \emph{arXiv preprint arXiv:1904.10509}, 2019.

\bibitem[Collobert \& Weston(2008)Collobert and Weston]{collobert2008unified}
Ronan Collobert and Jason Weston.
\newblock A unified architecture for natural language processing: Deep neural
  networks with multitask learning.
\newblock In \emph{Proceedings of the 25th international conference on Machine
  learning}, pp.\  160--167. ACM, 2008.

\bibitem[Collobert et~al.(2011)Collobert, Weston, Bottou, Karlen, Kavukcuoglu,
  and Kuksa]{collobert2011natural}
Ronan Collobert, Jason Weston, L{\'e}on Bottou, Michael Karlen, Koray
  Kavukcuoglu, and Pavel Kuksa.
\newblock Natural language processing (almost) from scratch.
\newblock \emph{Journal of machine learning research}, 12\penalty0
  (Aug):\penalty0 2493--2537, 2011.

\bibitem[Cowan(1987)]{Cowan1987}
Ruth~Schwartz Cowan.
\newblock The consumption junction: A proposal for research strategies in the
  sociology of technology.
\newblock In Wiebe~E. Bijker, Thomas~P. Hughes, and Trevor~J. Pinch (eds.),
  \emph{The Social Construction of Technological Systems}, pp.\  261--280. MIT
  Press, Cambridge, MA, USA, 1987.

\bibitem[Dai \& Le(2015)Dai and Le]{dai2015semi}
Andrew~M Dai and Quoc~V Le.
\newblock Semi-supervised sequence learning.
\newblock In \emph{Advances in neural information processing systems}, pp.\
  3079--3087, 2015.

\bibitem[Dai et~al.(2019)Dai, Yang, Yang, Cohen, Carbonell, Le, and
  Salakhutdinov]{dai2019transformer}
Zihang Dai, Zhilin Yang, Yiming Yang, William~W Cohen, Jaime Carbonell, Quoc~V
  Le, and Ruslan Salakhutdinov.
\newblock Transformer-xl: Attentive language models beyond a fixed-length
  context.
\newblock \emph{arXiv preprint arXiv:1901.02860}, 2019.

\bibitem[Devlin et~al.(2018)Devlin, Chang, Lee, and Toutanova]{devlin2018bert}
Jacob Devlin, Ming-Wei Chang, Kenton Lee, and Kristina Toutanova.
\newblock Bert: Pre-training of deep bidirectional transformers for language
  understanding.
\newblock \emph{arXiv preprint arXiv:1810.04805}, 2018.

\bibitem[Duchi et~al.(2011)Duchi, Hazan, and Singer]{adagrad}
John Duchi, Elad Hazan, and Yoram Singer.
\newblock Adaptive subgradient methods for online learning and stochastic
  optimization.
\newblock \emph{Journal of Machine Learning Research}, 12\penalty0
  (Jul):\penalty0 2121--2159, 2011.

\bibitem[Dunn et~al.(2017)Dunn, Sagun, Higgins, Guney, Cirik, and
  Cho]{dunn2017searchqa}
Matthew Dunn, Levent Sagun, Mike Higgins, V~Ugur Guney, Volkan Cirik, and
  Kyunghyun Cho.
\newblock Searchqa: A new q\&a dataset augmented with context from a search
  engine.
\newblock \emph{arXiv preprint arXiv:1704.05179}, 2017.

\bibitem[Fan et~al.(2018)Fan, Lewis, and Dauphin]{fan2018hierarchical}
Angela Fan, Mike Lewis, and Yann Dauphin.
\newblock Hierarchical neural story generation.
\newblock \emph{arXiv preprint arXiv:1805.04833}, 2018.

\bibitem[Fan et~al.(2019)Fan, Jernite, Perez, Grangier, Weston, and
  Auli]{fan2019eli5}
Angela Fan, Yacine Jernite, Ethan Perez, David Grangier, Jason Weston, and
  Michael Auli.
\newblock Eli5: Long form question answering.
\newblock \emph{arXiv preprint arXiv:1907.09190}, 2019.

\bibitem[Ginsburg et~al.(2019)Ginsburg, Castonguay, Hrinchuk, Kuchaiev,
  Lavrukhin, Leary, Li, Nguyen, and Cohen]{novograd}
Boris Ginsburg, Patrice Castonguay, Oleksii Hrinchuk, Oleksii Kuchaiev, Vitaly
  Lavrukhin, Ryan Leary, Jason Li, Huyen Nguyen, and Jonathan~M Cohen.
\newblock Stochastic gradient methods with layer-wise adaptive moments for
  training of deep networks.
\newblock \emph{arXiv preprint arXiv:1905.11286}, 2019.

\bibitem[Goodfellow et~al.(2014)Goodfellow, Pouget-Abadie, Mirza, Xu,
  Warde-Farley, Ozair, Courville, and Bengio]{goodfellow2014generative}
Ian Goodfellow, Jean Pouget-Abadie, Mehdi Mirza, Bing Xu, David Warde-Farley,
  Sherjil Ozair, Aaron Courville, and Yoshua Bengio.
\newblock Generative adversarial nets.
\newblock In \emph{Advances in neural information processing systems}, pp.\
  2672--2680, 2014.

\bibitem[Grusky et~al.(2018)Grusky, Naaman, and Artzi]{N18-1065}
Max Grusky, Mor Naaman, and Yoav Artzi.
\newblock Newsroom: A dataset of 1.3 million summaries with diverse extractive
  strategies.
\newblock In \emph{Proceedings of the 2018 Conference of the North American
  Chapter of the Association for Computational Linguistics: Human Language
  Technologies}, pp.\  708--719, New Orleans, Louisiana, June 2018. Association
  for Computational Linguistics.
\newblock URL \url{http://aclweb.org/anthology/N18-1065}.

\bibitem[Hashimoto et~al.(2016)Hashimoto, Xiong, Tsuruoka, and
  Socher]{hashimoto2016joint}
Kazuma Hashimoto, Caiming Xiong, Yoshimasa Tsuruoka, and Richard Socher.
\newblock A joint many-task model: Growing a neural network for multiple nlp
  tasks.
\newblock \emph{arXiv preprint arXiv:1611.01587}, 2016.

\bibitem[He et~al.(2016)He, Zhang, Ren, and Sun]{he2016deep}
Kaiming He, Xiangyu Zhang, Shaoqing Ren, and Jian Sun.
\newblock Deep residual learning for image recognition.
\newblock In \emph{Proceedings of the IEEE conference on computer vision and
  pattern recognition}, pp.\  770--778, 2016.

\bibitem[Hermann et~al.(2015)Hermann, Kocisky, Grefenstette, Espeholt, Kay,
  Suleyman, and Blunsom]{hermann2015teaching}
Karl~Moritz Hermann, Tomas Kocisky, Edward Grefenstette, Lasse Espeholt, Will
  Kay, Mustafa Suleyman, and Phil Blunsom.
\newblock Teaching machines to read and comprehend.
\newblock In \emph{Advances in neural information processing systems}, pp.\
  1693--1701, 2015.

\bibitem[Holtzman et~al.(2019)Holtzman, Buys, Forbes, and Choi]{nucleus}
Ari Holtzman, Jan Buys, Maxwell Forbes, and Yejin Choi.
\newblock The curious case of neural text degeneration.
\newblock \emph{arXiv preprint arXiv:1904.09751}, 2019.

\bibitem[Howard \& Ruder(2018)Howard and Ruder]{howard2018universal}
Jeremy Howard and Sebastian Ruder.
\newblock Universal language model fine-tuning for text classification.
\newblock \emph{arXiv preprint arXiv:1801.06146}, 2018.

\bibitem[Inan et~al.(2016)Inan, Khosravi, and Socher]{inan}
Hakan Inan, Khashayar Khosravi, and Richard Socher.
\newblock Tying word vectors and word classifiers: A loss framework for
  language modeling.
\newblock \emph{arXiv preprint arXiv:1611.01462}, 2016.

\bibitem[Johnson et~al.(2017)Johnson, Schuster, Le, Krikun, Wu, Chen, Thorat,
  Vi{\'e}gas, Wattenberg, Corrado, et~al.]{johnson2017google}
Melvin Johnson, Mike Schuster, Quoc~V Le, Maxim Krikun, Yonghui Wu, Zhifeng
  Chen, Nikhil Thorat, Fernanda Vi{\'e}gas, Martin Wattenberg, Greg Corrado,
  et~al.
\newblock Google’s multilingual neural machine translation system: Enabling
  zero-shot translation.
\newblock \emph{Transactions of the Association for Computational Linguistics},
  5:\penalty0 339--351, 2017.

\bibitem[Joshi et~al.(2017)Joshi, Choi, Weld, and
  Zettlemoyer]{joshi2017triviaqa}
Mandar Joshi, Eunsol Choi, Daniel~S Weld, and Luke Zettlemoyer.
\newblock Triviaqa: A large scale distantly supervised challenge dataset for
  reading comprehension.
\newblock \emph{arXiv preprint arXiv:1705.03551}, 2017.

\bibitem[Kaiser \& Moreno(2012)Kaiser and Moreno]{KaiserM2012}
David Kaiser and Jonathan Moreno.
\newblock Self-censorship is not enough.
\newblock \emph{Nature}, 492\penalty0 (7429):\penalty0 345--347, December 2012.
\newblock \doi{10.1038/492345a}.

\bibitem[Kaiser et~al.(2017)Kaiser, Gomez, Shazeer, Vaswani, Parmar, Jones, and
  Uszkoreit]{kaiser2017one}
Lukasz Kaiser, Aidan~N Gomez, Noam Shazeer, Ashish Vaswani, Niki Parmar, Llion
  Jones, and Jakob Uszkoreit.
\newblock One model to learn them all.
\newblock \emph{arXiv preprint arXiv:1706.05137}, 2017.

\bibitem[Kaiser et~al.(2018)Kaiser, Roy, Vaswani, Parmar, Bengio, Uszkoreit,
  and Shazeer]{kaiser2018fast}
{\L}ukasz Kaiser, Aurko Roy, Ashish Vaswani, Niki Parmar, Samy Bengio, Jakob
  Uszkoreit, and Noam Shazeer.
\newblock Fast decoding in sequence models using discrete latent variables.
\newblock \emph{arXiv preprint arXiv:1803.03382}, 2018.

\bibitem[Keskar et~al.(2019)Keskar, McCann, Xiong, and
  Socher]{keskar2019unifying}
Nitish~Shirish Keskar, Bryan McCann, Caiming Xiong, and Richard Socher.
\newblock Unifying question answering and text classification via span
  extraction.
\newblock \emph{arXiv preprint arXiv:1904.09286}, 2019.

\bibitem[Kingma \& Ba(2014)Kingma and Ba]{adam}
Diederik~P Kingma and Jimmy Ba.
\newblock Adam: A method for stochastic optimization.
\newblock \emph{arXiv preprint arXiv:1412.6980}, 2014.

\bibitem[Kingma \& Welling(2013)Kingma and Welling]{kingma2013auto}
Diederik~P Kingma and Max Welling.
\newblock Auto-encoding variational bayes.
\newblock \emph{arXiv preprint arXiv:1312.6114}, 2013.

\bibitem[Kiros et~al.(2015)Kiros, Zhu, Salakhutdinov, Zemel, Urtasun, Torralba,
  and Fidler]{kiros2015skip}
Ryan Kiros, Yukun Zhu, Ruslan~R Salakhutdinov, Richard Zemel, Raquel Urtasun,
  Antonio Torralba, and Sanja Fidler.
\newblock Skip-thought vectors.
\newblock In \emph{Advances in neural information processing systems}, pp.\
  3294--3302, 2015.

\bibitem[Kobus et~al.(2016)Kobus, Crego, and Senellart]{kobus2016domain}
Catherine Kobus, Josep Crego, and Jean Senellart.
\newblock Domain control for neural machine translation.
\newblock \emph{arXiv preprint arXiv:1612.06140}, 2016.

\bibitem[Kry{\'s}ci{\'n}ski et~al.(2019)Kry{\'s}ci{\'n}ski, Keskar, McCann,
  Xiong, and Socher]{kryscinski2019neural}
Wojciech Kry{\'s}ci{\'n}ski, Nitish~Shirish Keskar, Bryan McCann, Caiming
  Xiong, and Richard Socher.
\newblock Neural text summarization: A critical evaluation.
\newblock \emph{arXiv preprint arXiv:1908.08960}, 2019.

\bibitem[Kwiatkowski et~al.(2019)Kwiatkowski, Palomaki, Redfield, Collins,
  Parikh, Alberti, Epstein, Polosukhin, Devlin, Lee,
  et~al.]{kwiatkowski2019natural}
Tom Kwiatkowski, Jennimaria Palomaki, Olivia Redfield, Michael Collins, Ankur
  Parikh, Chris Alberti, Danielle Epstein, Illia Polosukhin, Jacob Devlin,
  Kenton Lee, et~al.
\newblock Natural questions: a benchmark for question answering research.
\newblock \emph{Transactions of the Association for Computational Linguistics},
  7:\penalty0 453--466, 2019.

\bibitem[Lample \& Conneau(2019)Lample and Conneau]{lample2019cross}
Guillaume Lample and Alexis Conneau.
\newblock Cross-lingual language model pretraining.
\newblock \emph{arXiv preprint arXiv:1901.07291}, 2019.

\bibitem[Lample et~al.(2019)Lample, Sablayrolles, Ranzato, Denoyer, and
  J{\'e}gou]{lample2019large}
Guillaume Lample, Alexandre Sablayrolles, Marc'Aurelio Ranzato, Ludovic
  Denoyer, and Herv{\'e} J{\'e}gou.
\newblock Large memory layers with product keys.
\newblock \emph{arXiv preprint arXiv:1907.05242}, 2019.

\bibitem[Levesque et~al.(2012)Levesque, Davis, and
  Morgenstern]{levesque2012winograd}
Hector Levesque, Ernest Davis, and Leora Morgenstern.
\newblock The winograd schema challenge.
\newblock In \emph{Thirteenth International Conference on the Principles of
  Knowledge Representation and Reasoning}, 2012.

\bibitem[Lewis et~al.(2019)Lewis, Denoyer, and Riedel]{lewis2019unsupervised}
Patrick Lewis, Ludovic Denoyer, and Sebastian Riedel.
\newblock Unsupervised question answering by cloze translation.
\newblock \emph{arXiv preprint arXiv:1906.04980}, 2019.

\bibitem[Luong et~al.(2015)Luong, Le, Sutskever, Vinyals, and
  Kaiser]{luong2015multi}
Minh-Thang Luong, Quoc~V Le, Ilya Sutskever, Oriol Vinyals, and Lukasz Kaiser.
\newblock Multi-task sequence to sequence learning.
\newblock \emph{arXiv preprint arXiv:1511.06114}, 2015.

\bibitem[McAuley et~al.(2015)McAuley, Targett, Shi, and Van
  Den~Hengel]{mcauley2015image}
Julian McAuley, Christopher Targett, Qinfeng Shi, and Anton Van Den~Hengel.
\newblock Image-based recommendations on styles and substitutes.
\newblock In \emph{Proceedings of the 38th International ACM SIGIR Conference
  on Research and Development in Information Retrieval}, pp.\  43--52. ACM,
  2015.

\bibitem[McCann et~al.(2017)McCann, Bradbury, Xiong, and
  Socher]{mccann2017learned}
Bryan McCann, James Bradbury, Caiming Xiong, and Richard Socher.
\newblock Learned in translation: Contextualized word vectors.
\newblock In \emph{Advances in Neural Information Processing Systems}, pp.\
  6294--6305, 2017.

\bibitem[McCann et~al.(2018)McCann, Keskar, Xiong, and
  Socher]{mccann2018natural}
Bryan McCann, Nitish~Shirish Keskar, Caiming Xiong, and Richard Socher.
\newblock The natural language decathlon: Multitask learning as question
  answering.
\newblock \emph{arXiv preprint arXiv:1806.08730}, 2018.

\bibitem[Merity et~al.(2017)Merity, Keskar, and Socher]{awdlstm}
Stephen Merity, Nitish~Shirish Keskar, and Richard Socher.
\newblock Regularizing and optimizing lstm language models.
\newblock \emph{arXiv preprint arXiv:1708.02182}, 2017.

\bibitem[Mikolov et~al.(2013)Mikolov, Sutskever, Chen, Corrado, and
  Dean]{mikolov2013distributed}
Tomas Mikolov, Ilya Sutskever, Kai Chen, Greg~S Corrado, and Jeff Dean.
\newblock Distributed representations of words and phrases and their
  compositionality.
\newblock In \emph{Advances in neural information processing systems}, pp.\
  3111--3119, 2013.

\bibitem[Mitchell et~al.(2019)Mitchell, Wu, Zaldivar, Barnes, Vasserman,
  Hutchinson, Spitzer, Raji, and Gebru]{MitchellWZBVHSRG2019}
Margaret Mitchell, Simone Wu, Andrew Zaldivar, Parker Barnes, Lucy Vasserman,
  Ben Hutchinson, Elena Spitzer, Inioluwa~Deborah Raji, and Timnit Gebru.
\newblock Model cards for model reporting.
\newblock In \emph{Proceedings of the Conference on Fairness, Accountability,
  and Transparency (FAT* '19)}, January 2019.
\newblock \doi{10.1145/3287560.3287596}.

\bibitem[Moryossef et~al.(2019)Moryossef, Aharoni, and
  Goldberg]{moryossef2019filling}
Amit Moryossef, Roee Aharoni, and Yoav Goldberg.
\newblock Filling gender \& number gaps in neural machine translation with
  black-box context injection.
\newblock \emph{arXiv preprint arXiv:1903.03467}, 2019.

\bibitem[Nair \& Hinton(2010)Nair and Hinton]{Nair2010RectifiedLU}
Vinod Nair and Geoffrey~E Hinton.
\newblock Rectified linear units improve restricted boltzmann machines.
\newblock In \emph{Proceedings of the 27th International Conference on Machine
  Learning (ICML-10)}, pp.\  807--814, 2010.

\bibitem[Nallapati et~al.(2016)Nallapati, Zhou, Gulcehre, Xiang,
  et~al.]{nallapati2016abstractive}
Ramesh Nallapati, Bowen Zhou, Caglar Gulcehre, Bing Xiang, et~al.
\newblock Abstractive text summarization using sequence-to-sequence rnns and
  beyond.
\newblock \emph{arXiv preprint arXiv:1602.06023}, 2016.

\bibitem[Peters et~al.(2018)Peters, Neumann, Iyyer, Gardner, Clark, Lee, and
  Zettlemoyer]{peters2018deep}
Matthew~E Peters, Mark Neumann, Mohit Iyyer, Matt Gardner, Christopher Clark,
  Kenton Lee, and Luke Zettlemoyer.
\newblock Deep contextualized word representations.
\newblock \emph{arXiv preprint arXiv:1802.05365}, 2018.

\bibitem[Pfaff(1979)]{pfaff1979constraints}
Carol~W Pfaff.
\newblock Constraints on language mixing: intrasentential code-switching and
  borrowing in spanish/english.
\newblock \emph{Language}, pp.\  291--318, 1979.

\bibitem[Poplack(1980)]{poplack1980sometimes}
Shana Poplack.
\newblock Sometimes i’ll start a sentence in spanish y termino en espanol:
  toward a typology of code-switching1.
\newblock \emph{Linguistics}, 18\penalty0 (7-8):\penalty0 581--618, 1980.

\bibitem[Press \& Wolf(2016)Press and Wolf]{press2016using}
Ofir Press and Lior Wolf.
\newblock Using the output embedding to improve language models.
\newblock \emph{arXiv preprint arXiv:1608.05859}, 2016.

\bibitem[Radford et~al.(2018)Radford, Narasimhan, Salimans, and
  Sutskever]{radford2018improving}
Alec Radford, Karthik Narasimhan, Tim Salimans, and Ilya Sutskever.
\newblock Improving language understanding by generative pre-training.
\newblock \emph{URL
  \texttt{https://s3-us-west-2.amazonaws.com/\\openai-assets/research-covers/langu\\ageunsupervised/language\_understand\\ing\_paper.pdf}},
  2018.

\bibitem[Radford et~al.(2019)Radford, Wu, Child, Luan, Amodei, and
  Sutskever]{radford2019language}
Alec Radford, Jeffrey Wu, Rewon Child, David Luan, Dario Amodei, and Ilya
  Sutskever.
\newblock Language models are unsupervised multitask learners.
\newblock \emph{URL
  \texttt{https://d4mucfpksywv.cloudfront.net\\/better-language-models/language\_mo\\dels\_are\_unsupervised\_multitask\_learn\\ers.pdf}},
  2019.

\bibitem[Rajani et~al.(2019)Rajani, McCann, Xiong, and
  Socher]{rajani2019explain}
Nazneen~Fatema Rajani, Bryan McCann, Caiming Xiong, and Richard Socher.
\newblock Explain yourself! leveraging language models for commonsense
  reasoning.
\newblock \emph{arXiv preprint arXiv:1906.02361}, 2019.

\bibitem[Rajpurkar et~al.(2016)Rajpurkar, Zhang, Lopyrev, and
  Liang]{rajpurkar2016squad}
Pranav Rajpurkar, Jian Zhang, Konstantin Lopyrev, and Percy Liang.
\newblock Squad: 100,000+ questions for machine comprehension of text.
\newblock \emph{arXiv preprint arXiv:1606.05250}, 2016.

\bibitem[Rush et~al.(2015)Rush, Chopra, and Weston]{rush2015neural}
Alexander~M Rush, Sumit Chopra, and Jason Weston.
\newblock A neural attention model for abstractive sentence summarization.
\newblock \emph{arXiv preprint arXiv:1509.00685}, 2015.

\bibitem[Sandhaus(2008)]{sandhaus2008new}
Evan Sandhaus.
\newblock The new york times annotated corpus.
\newblock \emph{Linguistic Data Consortium, Philadelphia}, 6\penalty0
  (12):\penalty0 e26752, 2008.

\bibitem[Scialom et~al.(2019)Scialom, Lamprier, Piwowarski, and
  Staiano]{scialom2019answers}
Thomas Scialom, Sylvain Lamprier, Benjamin Piwowarski, and Jacopo Staiano.
\newblock Answers unite! unsupervised metrics for reinforced summarization
  models.
\newblock \emph{arXiv preprint arXiv:1909.01610}, 2019.

\bibitem[See et~al.(2017)See, Liu, and Manning]{see2017get}
Abigail See, Peter~J Liu, and Christopher~D Manning.
\newblock Get to the point: Summarization with pointer-generator networks.
\newblock In \emph{Proceedings of the 55th Annual Meeting of the Association
  for Computational Linguistics (Volume 1: Long Papers)}, volume~1, pp.\
  1073--1083, 2017.

\bibitem[Sennrich et~al.(2015)Sennrich, Haddow, and Birch]{BPE}
Rico Sennrich, Barry Haddow, and Alexandra Birch.
\newblock Neural machine translation of rare words with subword units.
\newblock \emph{arXiv preprint arXiv:1508.07909}, 2015.

\bibitem[Shazeer \& Stern(2018)Shazeer and Stern]{adafactor}
Noam Shazeer and Mitchell Stern.
\newblock Adafactor: Adaptive learning rates with sublinear memory cost.
\newblock \emph{arXiv preprint arXiv:1804.04235}, 2018.

\bibitem[Stilgoe et~al.(2013)Stilgoe, Owen, and Macnaghten]{StilgoeOM2013}
Jack Stilgoe, Richard Owen, and Phil Macnaghten.
\newblock Developing a framework for responsible innovation.
\newblock \emph{Research Policy}, 42\penalty0 (9):\penalty0 1568--1580,
  November 2013.
\newblock \doi{10.1016/j.respol.2013.05.008}.

\bibitem[Sutskever et~al.(2014)Sutskever, Vinyals, and
  Le]{sutskever2014sequence}
Ilya Sutskever, Oriol Vinyals, and Quoc~V Le.
\newblock Sequence to sequence learning with neural networks.
\newblock In \emph{Advances in neural information processing systems}, pp.\
  3104--3112, 2014.

\bibitem[Trinh \& Le(2018)Trinh and Le]{trinh2018simple}
Trieu~H Trinh and Quoc~V Le.
\newblock A simple method for commonsense reasoning.
\newblock \emph{arXiv preprint arXiv:1806.02847}, 2018.

\bibitem[Trischler et~al.(2016)Trischler, Wang, Yuan, Harris, Sordoni, Bachman,
  and Suleman]{trischler2016newsqa}
Adam Trischler, Tong Wang, Xingdi Yuan, Justin Harris, Alessandro Sordoni,
  Philip Bachman, and Kaheer Suleman.
\newblock Newsqa: A machine comprehension dataset.
\newblock \emph{arXiv preprint arXiv:1611.09830}, 2016.

\bibitem[Varshney et~al.(2019)Varshney, Keskar, and Socher]{VarshneyKS2019}
Lav~R. Varshney, Nitish~Shirish Keskar, and Richard Socher.
\newblock Pretrained {AI} models: Performativity, mobility, and change,
  September 2019.
\newblock arXiv:1909.03290 [cs.CY].

\bibitem[Vaswani et~al.(2017)Vaswani, Shazeer, Parmar, Uszkoreit, Jones, Gomez,
  Kaiser, and Polosukhin]{vaswani2017attention}
Ashish Vaswani, Noam Shazeer, Niki Parmar, Jakob Uszkoreit, Llion Jones,
  Aidan~N Gomez, \L~ukasz Kaiser, and Illia Polosukhin.
\newblock Attention is all you need.
\newblock In I.~Guyon, U.~V. Luxburg, S.~Bengio, H.~Wallach, R.~Fergus,
  S.~Vishwanathan, and R.~Garnett (eds.), \emph{Advances in Neural Information
  Processing Systems 30}, pp.\  5998--6008. Curran Associates, Inc., 2017.
\newblock URL
  \url{http://papers.nips.cc/paper/7181-attention-is-all-you-need.pdf}.

\bibitem[Wang et~al.(2018)Wang, Singh, Michael, Hill, Levy, and
  Bowman]{wang2018glue}
Alex Wang, Amapreet Singh, Julian Michael, Felix Hill, Omer Levy, and Samuel~R
  Bowman.
\newblock Glue: A multi-task benchmark and analysis platform for natural
  language understanding.
\newblock \emph{arXiv preprint arXiv:1804.07461}, 2018.

\bibitem[Welleck et~al.(2019)Welleck, Kulikov, Roller, Dinan, Cho, and
  Weston]{unlikelihood}
Sean Welleck, Ilia Kulikov, Stephen Roller, Emily Dinan, Kyunghyun Cho, and
  Jason Weston.
\newblock Neural text generation with unlikelihood training.
\newblock \emph{arXiv preprint arXiv:1908.04319}, 2019.

\bibitem[Wu et~al.(2016)Wu, Schuster, Chen, Le, Norouzi, Macherey, Krikun, Cao,
  Gao, Macherey, et~al.]{wu2016google}
Yonghui Wu, Mike Schuster, Zhifeng Chen, Quoc~V Le, Mohammad Norouzi, Wolfgang
  Macherey, Maxim Krikun, Yuan Cao, Qin Gao, Klaus Macherey, et~al.
\newblock Google's neural machine translation system: Bridging the gap between
  human and machine translation.
\newblock \emph{arXiv preprint arXiv:1609.08144}, 2016.

\bibitem[Xenouleas et~al.(2019)Xenouleas, Malakasiotis, Apidianaki, and
  Androutsopoulos]{xenouleas2019sumqe}
Stratos Xenouleas, Prodromos Malakasiotis, Marianna Apidianaki, and Ion
  Androutsopoulos.
\newblock Sumqe: a bert-based summary quality estimation model.
\newblock \emph{arXiv preprint arXiv:1909.00578}, 2019.

\bibitem[Yang et~al.(2018)Yang, Qi, Zhang, Bengio, Cohen, Salakhutdinov, and
  Manning]{yang2018hotpotqa}
Zhilin Yang, Peng Qi, Saizheng Zhang, Yoshua Bengio, William~W Cohen, Ruslan
  Salakhutdinov, and Christopher~D Manning.
\newblock Hotpotqa: A dataset for diverse, explainable multi-hop question
  answering.
\newblock \emph{arXiv preprint arXiv:1809.09600}, 2018.

\bibitem[Zellers et~al.(2019)Zellers, Holtzman, Rashkin, Bisk, Farhadi,
  Roesner, and Choi]{zellers2019defending}
Rowan Zellers, Ari Holtzman, Hannah Rashkin, Yonatan Bisk, Ali Farhadi,
  Franziska Roesner, and Yejin Choi.
\newblock Defending against neural fake news.
\newblock \emph{arXiv preprint arXiv:1905.12616}, 2019.

\end{thebibliography}
